\pdfoutput=1

\documentclass[11pt]{article}

\usepackage[final]{EMNLP2023}
\usepackage{times}
\usepackage{latexsym}
\usepackage{multirow}
\usepackage{booktabs}
\usepackage{graphicx}
\usepackage{amssymb}
\usepackage{makecell}
\usepackage{amsmath, bm}
\usepackage{longtable}
\usepackage{subfigure}
\usepackage{caption}
\usepackage{enumerate}
\usepackage{enumitem}

\usepackage{algorithm}
\usepackage{algorithmic}
\usepackage{longtable}
\usepackage{xspace}
\usepackage{comment}

\usepackage{xcolor}
\usepackage{color}

\usepackage[T1]{fontenc}

\usepackage[utf8]{inputenc}

\usepackage{microtype}

\usepackage{inconsolata}

\newcommand{\ours}{{BANER}\xspace}

\newcommand{\nonetoken}{\textsc{O}\xspace}

\title{BANER: Boundary-Aware LLMs \\for Few-Shot Named Entity Recognition}

\author{Quanjiang Guo$^{1}$, Yihong Dong$^{1}$, Ling Tian$^{1}$, Zhao Kang$^{1}$\thanks{Corresponding Author} , Yu Zhang$^{2}$, Sijie Wang$^{3}$\\
        $^1$ University of Electronic Science and Technology of China, Chengdu, China \\ 
        $^{2}$ Harbin Institute of Technology, Shenzhen, China  \\
        $^{3}$ Nanyang Technological University, Singapore  \\
        \texttt{guochance1999@163.com}, \texttt{dongyihong8@163.com}, \texttt{lingtian@uestc.edu.cn}, \\
        \texttt{zkang@uestc.edu.cn}, \texttt{yuzhang2717@gmail.com}, \texttt{wang1679@e.ntu.edu.sg}}
\begin{document}
\maketitle

\begin{abstract}
Despite the recent success of two-stage prototypical networks in few-shot named entity recognition (NER), challenges such as over/under-detected false spans in the span detection stage and unaligned entity prototypes in the type classification stage persist.   Additionally, LLMs have not proven to be effective few-shot information extractors in general. 
In this paper, we propose an approach called \textbf{B}oundary-\textbf{A}ware LLMs for Few-Shot \textbf{N}amed \textbf{E}ntity \textbf{R}ecognition (\ours) to address these issues.
We introduce a \emph{boundary-aware contrastive learning strategy} to enhance the LLM's ability to perceive entity boundaries for generalized entity spans. Additionally, we utilize LoRAHub to align information from the target domain to the source domain, thereby enhancing adaptive cross-domain classification capabilities.
Extensive experiments across various benchmarks demonstrate that our \ours framework outperforms prior methods, validating its effectiveness. In particular, the proposed strategies demonstrate effectiveness across a range of LLM architectures.
\footnote{The code and data are released on 
\url{https://github.com/UESTC-GQJ/BANER}.}
\end{abstract}

\section{Introduction}
Named Entity Recognition (NER) is a fundamental task in Natural Language Processing 
(NLP) that aims to detect the entity spans of text and classify them into pre-defined set of entity types. When there are sufficient labeled data, deep learning-based methods~\cite{huang2015bidirectional,ma2016end,lample2016neural,chiu2016named} have achieved impressive performance. However, in practical applications, it is desirable to recognize new entity types that have not been seen in the source domain. It is time-consuming and labor-expensive to collect extra labeled data for these new types. Consequently, few-shot NER~\cite{fritzler-2019,yang-katiyar-2020-simple}, which involves identifying unseen entity types based on only a few labeled samples for each class (also known as \emph{support samples}) in the target domain, has attracted a lot of attention in recent years.

\begin{figure}[t!]
\centering
\includegraphics[width=0.48\textwidth]{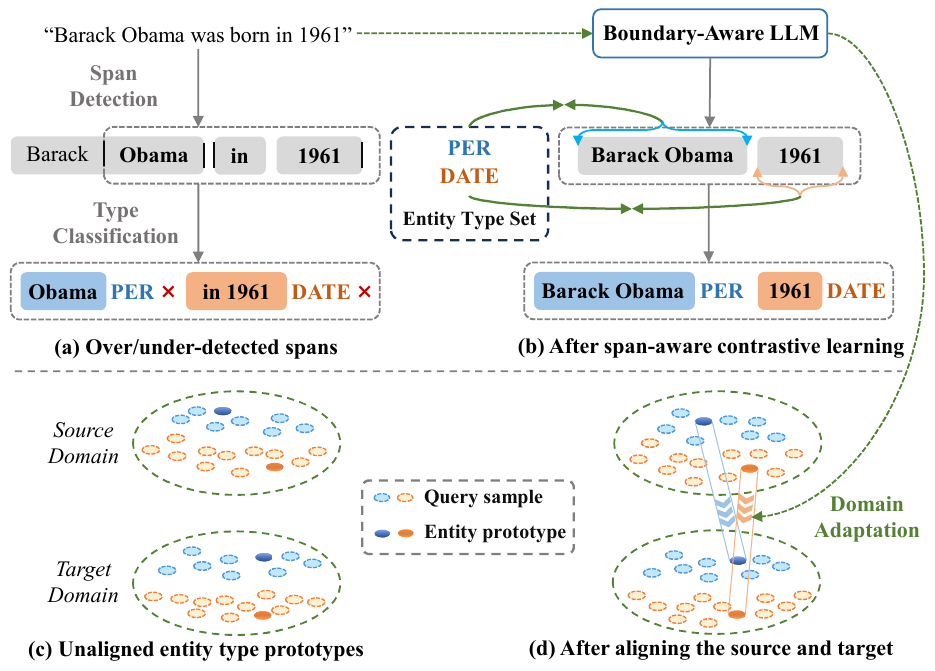}
\caption{(a) shows under/over-detected false spans, (b) shows correct spans obtained by adopting our boundary-aware LLM,
(c) shows unaligned entity type prototypes, (d) shows aligned prototypes obtained by our domain adaption strategy. 
}
\label{fig_introduction_comparision}
\vspace{-3mm}
\end{figure}
Previously end-to-end metric learning based methods~\cite{yang-katiyar-2020-simple, das-etal-2022-container} dominate the field of few-shot NER. These approaches are designed to learn the  intricate structure that includes both entity boundaries and entity types. However, their performance may degrade significantly when encountering a substantial domain gap. This degradation is primarily due to the challenge of understanding such complex structural information with only a few support examples for domain adaptation. Consequently, these methods often suffer from inadequate perception of boundary information, resulting in frequent misclassification of entity spans.
Though LLMs have made
remarkable success in various tasks, they have not proven to be effective few-shot information extractors in general \cite{ma2023large,zhang2024linkner}.

Recent works demonstrate that adopting two-stage prototypical networks~\cite{wang-2022-enhanced,ma-etal-2022-decomposed, li-etal-2023-type-aware} can be effective to address aforementioned issue, which decompose NER task into two distinct stages:   \emph{entity span detection} and \emph{entity type classification} tasks, executing each task sequentially.
Since decomposed methods only need to locate the spans of named entities and are class-agnostic in the first stage, they can identify more accurate entity boundaries and achieve better performance than end-to-end approaches.\looseness-1

While these two-stage prototypical methods have shown promising progress, they also present two additional challenges. Firstly, at the entity span detection stage, these decomposed approaches merely detect possible spans, often overlooking the boundary-related semantic information of named entities. For instance, following entity span detection, the sentence in Figure~\ref{fig_introduction_comparision}(a) illustrates that the span for ``Barack Obama'' is inadequately detected, resulting in ``Obama'' being identified while ``Barack'' is overlooked. Conversely, the span for ``1961'' is excessively detected as ``in 1961''.  These inaccuracies propagate errors into the subsequent entity type classification stage.

Secondly, in decomposed methods, prototypical networks aim to learn a type-related metric similarity function from test samples to classify entities based on their distance to prototypes. However, since the obtained prototypes are independently trained relative to the first stage, they may overlook entity type knowledge from the prior source domain. This can lead to difficulties in aligning the distribution of the same class across different domains. For example, in Figure~\ref{fig_introduction_comparision}(c), the entity types in the target domain exist independently of those in the source domain, leading to misaligned prototypes for the same entity type. This misalignment can severely impact the cross-domain performance of few-shot NER during the entity type classification stage.

To this end, we propose an approach called \textbf{B}oundary-\textbf{A}ware LLMs for Few-Shot \textbf{N}amed \textbf{E}ntity \textbf{R}ecognition (BANER). Our approach adopts the two-stage framework of the decomposed method but advances two steps further to effectively address the aforementioned challenges.
For \emph{entity span detection}, we design a boundary-aware contrastive learning strategy to reduce the gap between span embeddings of entities and their corresponding types using LLM. This strategy enhances the boundary perception capabilities of LLM, particularly for generalized entity spans. 
For \emph{entity type classification}, we draw upon domain adaptation principles to construct refined prototypes that retain and align entity type knowledge from the source domain. This approach involves joint pretraining in the source domain and adaptive alignment across diverse target domains within the same LLM framework, facilitated by LoRAHub~\cite{huang2023lorahub}. 

In summary, our contributions are as follows:

(1) We introduce a novel Few-Shot NER approach, \ours, which employs boundary-aware contrastive learning to enhance an LLM's ability to perceive entity boundaries. To our knowledge, this is the first integration of LLM with contrastive learning for few-shot NER tasks.

(2) Leveraging an LLM pretrained on the source domain, we utilize LoRAHub to align information from target domains to enhance adaptive cross-domain classification capabilities.

(3) Experimental results on multiple few-shot NER datasets demonstrate that \ours achieves superior performance compared to previous state-of-the-art two-stage decomposed methods. Furthermore, we validate the generalizability of our strategies across various LLM architectures.

\begin{figure*}[!htbp]
\centering
\includegraphics[width=0.95\textwidth]{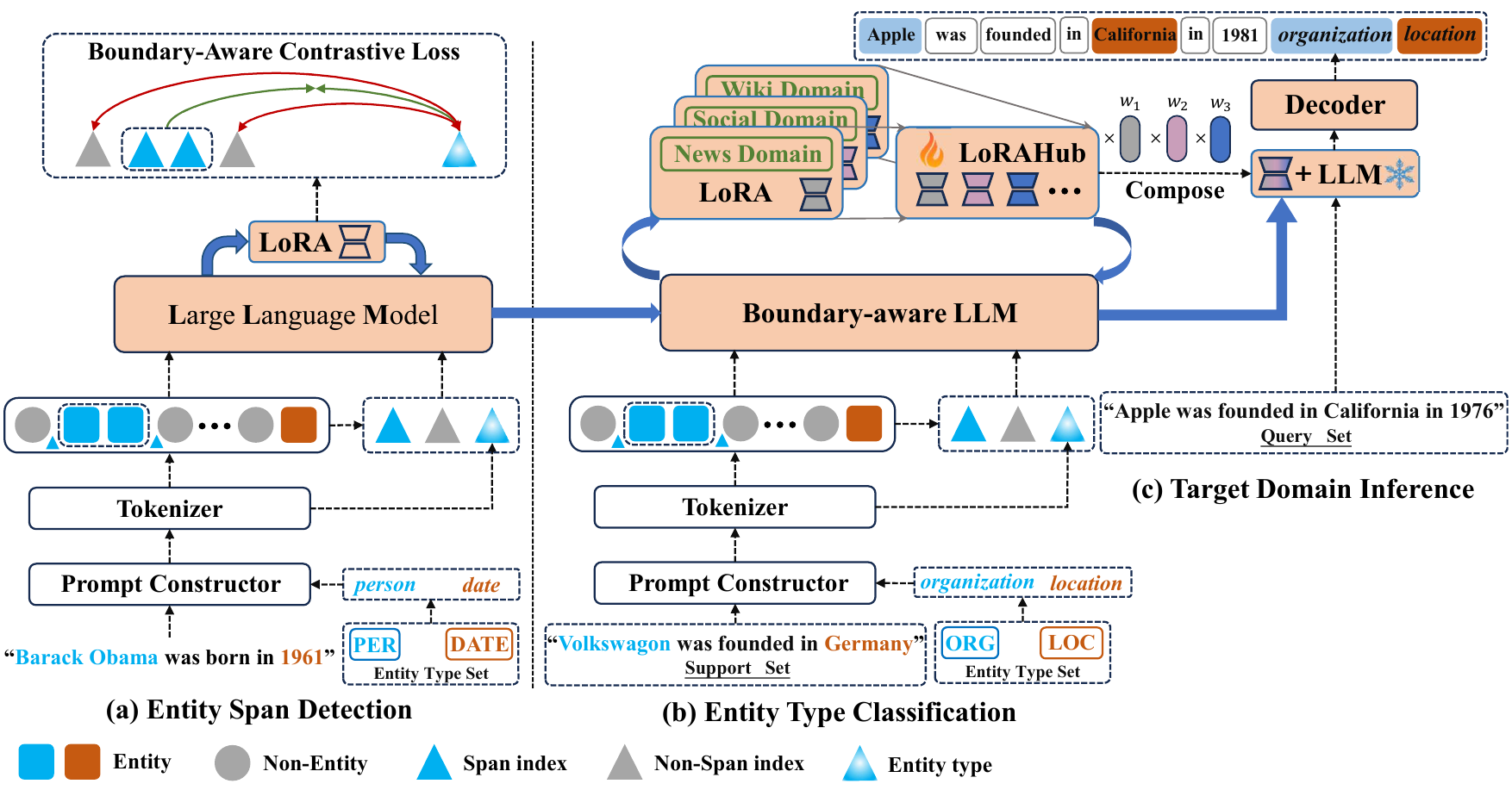}
\caption{Overall structure of the proposed \ours. (a) Entity span detection with pre-training in the source domain. (b) Entity type classification with fine-tuning in the support samples of target domain. (c) Inference on the query set of target domain.
}
\label{fig_framework}
\vspace{-0.2cm}
\end{figure*}

\section{Related Work}
\paragraph{Few-Shot NER}
Recently, few-shot named entity recognition (NER) has garnered considerable attention. Previous methods can be broadly categorized into two types: prompt-based and metric-based approaches. Prompt-based methods focus on leveraging the knowledge of pre-trained language models (LLMs) for NER through prompt learning techniques \cite{cui-etal-2021-template, ma-etal-2022-template, huang-etal-2022-copner, lee-etal-2022-good}. These methods utilize templates, prompts, or exemplary instances to effectively harness the pre-existing knowledge within LLMs.

With the rapid advancements in LLMs, there has been a surge in studies exploring direct prompting of LLMs for few-shot NER tasks \cite{wang2023gpt, xie2023empirical}. Additionally, there is emerging interest in straightforward instruction-tuning strategies \cite{zhou2024universalner}, or annotating raw data with LLMs to train task-specific foundational models for NER \cite{zhang2024linkner}. However, their performance often diminishes when tasked with generating text that adheres to specific structured formats and domains, which is crucial in few-shot NER scenarios. 

Metric-based methods, on the other hand, aim to learn a feature space with robust generalizability and classify test samples using nearest class prototypes \cite{snell_prototypical_2017, fritzler-2019, ji-etal-2022-shot, ma-etal-2022-decomposed} or neighboring samples \cite{yang-katiyar-2020-simple, das-etal-2022-container}. Nevertheless, the prototypical networks widely employed in these methods may fail to fully utilize entity type knowledge from the source domain during the type classification stage.

Moreover, recent research has focused on the two-stage architecture for few-shot named entity recognition (NER) \cite{shen-etal-2021-locate, wang2021learning-from-description, zhang2022exploring, wang-2022-enhanced, ma-etal-2022-decomposed, li-etal-2023-type-aware}, where the task is decomposed into entity span detection and entity typing subtasks. These methods excel in learning entity boundary information under data-limited conditions and often achieve superior performance. However, they may encounter challenges such as over/under-detection of false entity spans during the span detection stage.


\paragraph{Contrastive Learning and Domain Adaptation}
Due to the robust generalization capabilities of contrastive learning, recent methods \cite{das-etal-2022-container, huang-etal-2022-copner} have adopted this approach for few-shot NER, employing contrastive losses between tokens or between tokens and prompts. However, these methods are end-to-end approaches and therefore inherently lack the ability to effectively learn entity boundary information. In contrast, our approach is decomposed, and our boundary-aware contrastive loss is designed between the span embeddings of entities and their corresponding types within the LLM framework. This method enables the learning of a span-aware feature space in LLMs, facilitating accurate boundary detection.

Domain adaptation tackles the challenge of dataset shift between source and target domains, particularly when only a few samples are available in the target domain. When labels in the target domain are scarce, the problem transitions into a semi-supervised scenario. Traditional approaches combine source and target data to enhance model training \cite{zhang2021semi, zhang2024tpn}. In the context of type classification adaptation using LLMs, fine-tuning remains the predominant method \cite{grangier2022trade, guo2021bertweetfr, buonocore2023localizing}. Alternatively, strategies involve expanding the LLM's vocabulary with domain-specific tokens \cite{sachidananda2021efficient, zhu2024fcds} or employing adversarial adaptation techniques such as knowledge distillation \cite{rietzler2020adapt} or supervised fine-tuning \cite{ryu2022knowledge, zhang2024question}. In contrast, our approach leverages LoRAHub to dynamically align information from the target domain with that of the source domain.



\section{Methodology}

Figure~\ref{fig_framework} depicts  the overall framework of our \ours. Like other two-stage methods, it comprises \emph{entity span detection} and \emph{entity type classification}. Notably, our approach incorporates boundary-aware contrastive learning and adaptive domain alignment strategies at these respective stages.

\paragraph{Task Formulation}
Given a sequence $X=\{x_i\}^{L}_{i=1}$ with $L$ tokens, NER aims to assign each token $x_i$ to its corresponding label $y_i \in Y \cup  \nonetoken $, 
where $Y$ is the pre-defined entity type set and \nonetoken denotes non-entities. For few-shot NER, the NER model is first pretrained on data-sufficient source domain(s) $\mathcal{D}_{s}=\{({S}_{s},{Q}_{s},Y_{s})\}$ and then fine-tuned in target domain(s) $\mathcal{D}_{t}=\{({S}_{t},{Q}_{t},Y_{t})\}$ with only a few labeled samples. We adhere to the standard $N$-way $K$-shot setting as outlined in ~\cite{ding-etal-2021-nerd}, where ${S}_{s/t}=\{(x_i,y_i)\}^{N \times K}_{i=1}$ denotes the support set, ${Q}_{s/t}=\{(x_j,y_j)\}^{N' \times K'}_{j=1}$ denotes the query set, $|{Y}_{s}|=N$ and $|{Y}_{t}|=N'$. Our task is to recognize entities in the query set ${Q}_{t}$ from the target domain after adapting the model using its support set ${S}_{t}$.
It is noteworthy that $Y_{s}$ and $Y_{t}$ exhibit little to no overlap.

\subsection{Entity Span Detection}
\subsubsection{Prompt Representation}\label{section_span_detetor}
Formally, we denote the LLM as $f_\mathrm{LLM}$ 
 and input instruction as $I$. The output (generated) token sequence is denoted as 
$Y=f_\mathrm{LLM}(X)=\{y_i\}^{L}_{i=1}$.
For the classic auto-regressive generative model, the sampling probability of the model generating $Y$ is formalized as follows:

\begin{equation}
\small
\mathbb{P}(Y \mid I, X) = \prod_{t=1}^{L} \mathbb{P}(y_t \mid I, X, y_{<t}),
\end{equation}
where $y_t$ is the $t$-th token of the $y$, $y_{<t}$ represents the tokens before $y_t$.
Utilizing generative language models for information extraction typically involves providing a prompt as input and generating results according to a specified format.
In \ours, we adopt the default template for LoRA fine-tuning\footnote{\url{https://github.com/tatsu-lab/stanford_alpaca}}. The prompt is fed into the LLM to perform entity span detection. An example of such a prompt is illustrated in Figure~\ref{fig_prompt1} in Appendix~\ref{prompt}.

According to the LLM's token generation rule, the objective loss for auto-regressively generating $Y$ is as follows:
\begin{equation}
\small
\mathcal{L}_g = -\sum_{(X,y)\in \mathcal{D}_{s}} \sum_{t=1}^{L} \log \mathbb{P}_{\theta+\theta_{L}}(y_t \mid I, X, y_{<t}),
\end{equation}
where $\theta$ is the original parameters of LLM, $\theta_{L}$ is the LoRA parameters. Note that we only update LoRA parameters during the training process.

\subsubsection{Boundary-Aware Contrastive Learning}\label{section_span_detetor}
We enumerate all $m$ spans $S=\{s_1, s_2, . . . , s_m\}$ 
for sequence $X$. 
For example, for sentence ``Barack Obama was born in 1961'', span indices (begin, end) of two entities are $\{(0,2),(5,6)\}$. We use $b_i$ and $e_i$ to denote the begin- and end- index representation of the span $s_i$ in constructed prompt, respectively. 

To enhance the LLM's ability to perceive entity boundaries, we employ the concept of contrastive learning \cite{NEURIPS2020_d89a66c7_supervised_contrastive_learning}. We utilize two types of boundary-aware index representations, as illustrated in  Figure~\ref{fig_framework}(a), to construct positive and negative samples for each entity mention and its corresponding entity type. Specifically, the positive sample $\mathrm{pos_i}$ of entity span is calculated by  concatenating  $h_{b_i}$ and $h_{{e_i}-1}$ as 
$\mathrm{pos_i}=[h_{b_i}, h_{{e_i}-1}]$,
where $h_{(\cdot)}=\mathrm{embedding}(\cdot)$ is the pre-trained tokenizer of LlaMA-2-7B. The negative sample $\mathrm{neg}_i$ of entity boundary is $\mathrm{neg}_i = [h_{{b_i}-1},h_{{b_i}-2},h_{e_i},h_{{e_i}+1}]$.
The original entity type representation $o$ is the (begin, end) indices of entity type from constructed prompt in the same way. 

To learn better boundary-aware feature space, we extract entity type embedding $e_{o}$, entity token embedding $e_{\mathrm{pos}_i}$ and $e_{\mathrm{neg}_i}$, from outputs $H \in \mathbb{R}^{B \times L \times D}$ of 25th hidden states layer in LlaMA-2, where $B$ is the batch size and $D$ is the hidden dimension. The calculation formula are:
\begin{equation}
\small
e_{o_i} = \mathrm{gather}(H, o_i) \in \mathbb{R}^{B \times 1 \times D},
\end{equation}
\begin{equation}
\small
e_{\mathrm{pos}_i} = \mathrm{gather}(H, \mathrm{pos}_i) \in \mathbb{R}^{B \times 2 \times D},
\end{equation}
\begin{equation}
\small
e_{\mathrm{neg}_i} = \mathrm{gather}(H, \mathrm{neg}_i) \in \mathbb{R}^{B \times 4 \times D},
\end{equation}
where $\mathrm{gather}()$ is a tensor operation commonly used in deep learning frameworks (e.g., PyTorch), which allows for the selection and extraction of specific elements from a higher-dimensional tensor $H$ based on specified indices. Then, we can calculate the boundary-aware contrastive loss by:
\begin{equation}
\small
\mathcal{L}_{\mathrm{cl}} = - \frac{1}{B} \sum_{i=1}^B \log\left( \sigma({\mathrm{sim}(o,\mathrm{pos}_i)} - {\mathrm{sim}(o,\mathrm{neg}_i)}) \right),
\end{equation}
\begin{equation}
\small
{\mathrm{sim}(o,\mathrm{pos}_i)} = \sum_{i=1}^m (\frac{e_o}{\|e_o\|_2} \cdot \frac{e_{\mathrm{pos}_i}}{\|e_{\mathrm{pos}_i}\|_2}) \in \mathbb{R}^B ,
\end{equation}
\begin{equation}
\small
{\mathrm{sim}(o,\mathrm{neg}_i)} = \sum_{i=1}^m (\frac{e_o}{\|e_o\|_2} \cdot \frac{e_{\mathrm{neg}_i}}{\|e_{\mathrm{neg}_i}\|_2}) \in \mathbb{R}^B ,
\end{equation}
where $\sigma()$ is the sigmoid function.

\subsubsection{LLM Fine-Tuning}\label{section_span_detetor}
We introduce instruction tuning to effectively and efficiently align the LLM with the span detection task. Following the standard supervised fine-tuning method, we minimize the auto-regressive loss calculated between the ground truth and the LLM output. In our approach, we mask the loss positions corresponding to the prompt part. Specific prompt formats, task-specific instructions, and ground truth details are provided in the Appendix~\ref{prompt}. However, directly fine-tuning the entire model can be computationally intensive and time-consuming. To address this, we propose a lightweight fine-tuning strategy using LoRA. This method involves freezing the pre-trained model parameters and introducing trainable rank decomposition matrices into each layer of the Transformer architecture. This approach facilitates lightweight fine-tuning while reducing GPU memory consumption. The final learning objective is computed as follows:

\begin{equation}
\small
\mathcal{L}_\mathrm{span} = \min_{\theta_{L_1}} ({\mathcal{L}_{g} + \lambda \mathcal{L}_\mathrm{cl}}),
\end{equation}
where $\theta_{L_1}$ is the LoRA parameters at the span detection stage and $\lambda$ is set to 0.001.

\subsection{Entity Type Classification}
Subsequently, we assign a specific entity class to each span identified during the entity span detection stage.
\subsubsection{Prompt and Prototype Representation}
As previously mentioned, a predefined (candidate) list of entity types must be input as the schema into the LLM to trigger type generation. Figure~\ref{fig_prompt2} in Appendix~\ref{prompt} illustrates an example of the prompt used for this stage. Using this prompt, the model constructs a prototype for each given entity type, which is then used to assign the correct type to each detected entity span.

To achieve this, we construct prototypical networks (ProtoNet) as the backbone, utilizing LoRA tuning across different domains. To leverage the knowledge from support examples in the target domain and align it with the source domain, we propose enhancing ProtoNet on the LLM with domain adaptation. This approach aims to create a more representative embedding space where text spans from different entity classes are more distinguishable.

Let $S_k=\{z_1, z_2, \ldots, z_n\}$ denote the set of entity type spans in the constructed prompt, which is contained in a given support set $S_t$ belonging to the entity class $y_k \in Y$. We compute the prototype $p_k$ for each $y_k$ by averaging the span representations of all $z_i \in S_k$:
\begin{equation}
\small
p_k(S_t) = \frac{1}{|S_k|}\sum_{i=1}^{|S_k|}z_i.
\label{eq:proto}
\end{equation}
\subsubsection{LoRA Tuning across Different Domains}

Given a training episode $\mathcal{D}_{t}$, we first utilize the support set ${S}_{t}$ to compute prototypes for all entity classes in ${Y}_{t}$ using Eq. \ref{eq:proto}. Subsequently, for each span $s_i$ in the query set ${Q}_{t}$, we calculate the probability that $s_i$ belongs to an entity class $y_k$ based on the distance between its span representation and the prototype of $y_k$:
\begin{equation}
\small
\mathbb{P}\left(y_{k} ; z_i\right)=\frac{exp \left\{-d\left(p_{k}\left(\mathcal{S}_{t}\right), s_i\right)\right\}}{\sum_{y_{i} \in Y} exp \left\{-d\left(p_{i}\left(\mathcal{S}_{t}\right), s_i\right)\right\}},
\end{equation}
To compute the distance function $d\left( \cdot, \cdot \right)$, we define it as follows:
\begin{equation}
\small
d\left(p_{k/i}\left(\mathcal{S}_{t}\right), s_i\right)= \frac{p_{k/i}\left(\mathcal{S}_{t}\right)}{\|p_{k/i}\left(\mathcal{S}_{t}\right)\|_2} \cdot \frac{ s_i}{\| s_i\|_2}.
\end{equation}
Our goal is to minimize the cross-entropy loss for each LoRA module in its corresponding target domain:
\begin{equation}
\small
\mathcal{L}_{t_{i}} = \min_{\theta_{L_2}} \left( - \sum_{z_i \in Q_t} \log \mathbb{P}_{\theta+\theta_{L_2}} \left( y_{k} ; z_i \right) \right),
\end{equation}
where $\theta_{L_2}$ is the LoRA parameters at the type classification stage.
\subsubsection{Composition of LoRA Modules}
As depicted in Figure~\ref{fig_framework}(b), we initially fine-tuned LoRA modules across various target domains. Specifically, for $M$ distinct domains, we fine-tune $M$ separate LoRA modules, each denoted as $m_i$ for the domain $\mathcal{D}_{t_i} \in \mathcal{D}_{t}$.
Each $m_i$ can be defined as the product $A_iB_i$, where $A_i \in \mathbb{R}^{d \times r}$ and $B_i \in \mathbb{R}^{r \times k}$ are trainable low-rank matrices, with the rank $r$ being significantly smaller than the dimensions $d$ and $k$. The combined LoRA module $\hat{m}$ can be obtained by:
\begin{equation}
\small
\hat{m} = (w_1A_1 \cdots + w_NA_N)(w_1B_1 + \cdots + w_NB_N).
\end{equation}
To find the optimal $w$, the optimization process is guided by the cross-entropy loss to identify the best set ${w_1, w_2, \cdots , w_N}$ that minimizes the loss $\mathcal{L}_{t_i}$ on the target domain. Additionally, we incorporate L1~regularization to penalize the sum of the absolute values of $w$, helping to prevent extreme values. Consequently, the final objective of LoRAHub is to minimize $\mathcal{L}_{t_i} + \alpha \cdot \sum_{i=1}^{N} |w_i|$, where $\alpha$ serves as a hyperparameter.

\subsection{Target Domain Inference}
As illustrated in Figure~\ref{fig_framework}(c), during target domain inference, we first extract candidate spans from query sentences and then classify these spans into specific entity types to obtain the final results. After training the LLM with boundary-aware contrastive learning, we generate candidate entity spans from a given sentence $X$ as follows: 
\begin{equation}
\small
P(S|X;\theta+{\theta_{L_1}}) = \prod_{i=1}^{N} P(y_i | y_{<t}, X; \theta+{\theta_{L_1}}).
\end{equation}
Next, we obtain the candidate span set $S_\mathrm{span}$, which includes all potential spans to be assigned entity types during the entity type classification stage. For these candidate spans, the entity types are classified as follows: 
\begin{equation}
\small
P(C|X, S;\theta+{\theta_{L_2}}) = \prod_{i=1}^{N} P(y_i | y_{<t}, X, S; \theta+{\theta_{L_2}}).
\end{equation}
Finally, we combine the results of span detection and type classification to determine the predicted labels for a sentence $X$ as follows:
\begin{equation}
\small
P(Y|S, C;\hat{\theta}) = P(S|X;\theta+{\theta_{L_1}}) \cdot P(C|X, S;\theta+{\theta_{L_2}}).
\end{equation}

\begin{table}[htbp]
    \centering
    \small
    \resizebox{\columnwidth}{!}{
        \begin{tabular}{ccccc}
        \toprule
           \textbf{Dataset}  & \textbf{Domain} & \textbf{\# Sentences}  & \textbf{\# Entities} & \textbf{\# Classes} \\
           \cmidrule(r){1-1} \cmidrule(r){2-5} 
           Few-NERD  & Wikipedia & 188k  &491k & 66\\
           OntoNotes & General & 76k &104k & 18\\
           I2B2 & Medical & 140k  &29k & 23\\
           CoNLL & News & 20k  &35k & 4\\
           WNUT & Social & 5k &3k & 6\\
           GUM & Wiki & 3k  &6k & 11\\
         \bottomrule
        \end{tabular}
    }
    \caption{Statistics of Datasets}
    \vspace{-3mm}
    \label{tab:dataset-statistic}
\end{table}

\section{Experiments}
\subsection{Experimental Setups}
\subsubsection{Datasets}
\paragraph{Few-NERD~\footnote{\url{https://ningding97.github.io/fewnerd/}} \cite{ding-etal-2021-nerd}} It is the largest few-shot NER dataset containing 66 fine-grained entity types across 8 coarse-grained categories. 
Two tasks are considered for this dataset: (1) \textbf{intra}, where all entities in the train/dev/test splits belong to different coarse-grained types, and (2) \textbf{inter}, where the train/dev/test splits may share coarse-grained types but have mutually exclusive fine-grained entity types.
\paragraph{Cross-Dataset} To evaluate cross domain adaption, we follow \citet{das-etal-2022-container}  and take OntoNotes 5.0 (General)~\cite{weischedel2013ontonotes} as our source domain, and evaluate few-shot domain adaptation performances on I2B2'14 (Medical)~\cite{stubbs-2015-i2b2-2014}, CoNLL'03 (News)~\cite{sang2003conll2003}, WNUT'17 (Social)~\cite{derczynski2017wnut17}, and GUM (Wiki)~\cite{zeldes2017gum} datasets.

The statistics of datasets are shown in Table \ref{tab:dataset-statistic}.

\subsubsection{Baselines}
We compare our proposed \ours with the \emph{one-stage} and \emph{two-stage} types. The \emph{one-stage} baselines include \textbf{ProtoBERT}~\cite{snell_prototypical_2017}, \textbf{NNShot}~\cite{wiseman2019label}, \textbf{StructShot}~\cite{yang-katiyar-2020-simple}, \textbf{CONTaiNER}~\cite{das-etal-2022-container} and \textbf{MANNER}~\cite{fang2023manner}. The \emph{two-stage} baselines include: \textbf{ESD}~\cite{wang-2022-enhanced}, \textbf{DecomposedMetaNER}~\cite{ma-etal-2022-decomposed}, \textbf{TadNER}~\cite{li-etal-2023-type-aware}, \textbf{TSFNER}~\cite{ji-kong-2024-novel-three}, and \textbf{BDCP}~\cite{xue2024robust}.

\subsubsection{Evaluation Details}
\paragraph{Evaluation on Few-NERD} 
Following the methodology of \citet{ma-etal-2022-decomposed}, we adopt the episode-level evaluation approach. Each episode consists of a support set and a query set, structured in the N-way K-shot format. During evaluation, our model trained on the source domain predicts on the query set using information from the support set. To ensure fairness in comparisons, we compute the Micro F1 score based on episode data processed according to \citet{ding-etal-2021-nerd}. Results are reported as the mean F1 score ± standard deviation across 5 random seeds.

\paragraph{Evaluation on Cross-Dataset} 
\citet{yang-katiyar-2020-simple} points out the limitation that sampling test episodes may not accurately reflect real-world performance due to varying data distributions. They advocate for sampling support sets and subsequently evaluating models on the original test set. Each support set consists of K examples for each label. The final Micro F1 scores and standard deviations are calculated based on different sampled support sets. Following \citet{yang-katiyar-2020-simple} and \citet{das-etal-2022-container}, we adopt this evaluation schema specifically for \textbf{cross-domain} settings. To ensure fair comparisons, we employ the support sets sampled according to the methodology proposed by \citet{das-etal-2022-container}\footnote{\url{https://github.com/psunlpgroup/CONTaiNER}}.

\begin{table}[htb]
    \centering
    \small
    \resizebox{\columnwidth}{!}{
        \begin{tabular}{ccc}
        \toprule
           \textbf{Parameters}  & \textbf{Value} & \textbf{\# Comment} \\
           \midrule
           temperature & 0 & control the randomness of generation\\
           top\_p & 1 & determine the cumulative probability for nucleus sampling \\
           top\_k & 65536 & limit the number of highest probability tokens considered \\
           num\_beams & 4 & set the number of beams for beam search  \\
           max\_new\_tokens & 128 & define the maximum number of tokens to generate \\
         \bottomrule
        \end{tabular}
    }
    \caption{Main parameters in inference.}
    \vspace{-3mm}
    \label{tab:hyper-parameters}
\end{table}

\subsubsection{Implementation Details}
To construct \ours, we utilize LLaMA-2-7B as the pre-trained LLM backbone with FP16 precision and employ LoRA for prompt-tuning and model inference. During source domain training, we optimize using AdamW \cite{loshchilov2018decoupled} with a learning rate of \( 3 \times 10^{-4} \), a batch size of 1, and training over five epochs with a micro batch size of one. The cutoff length is set to 256, and no validation set is used (i.e., \( \text{val\_set\_size} = 0 \)). For LoRA, we set \( r = 32 \), \( \alpha = 16 \), and a dropout rate of 0.05. Distributed Data Parallel (DDP) is not employed for parameter search during training.

For target domain inference, Table \ref{tab:hyper-parameters} outlines the key parameters used in result generation. To ensure the robustness of generative language model outputs, our method incorporates task-specific instructions as inputs for entity span detection and type classification. Implementation is carried out using PyTorch 1.9.0\footnote{\url{https://pytorch.org/}} and executed on two Tesla A800-80G GPUs.

\begin{table*}[ht]
    \centering
    \setlength{\tabcolsep}{1.2mm}
    \resizebox{\linewidth}{!}{
    \begin{tabular}{l l  ccccc  ccccc}
    \toprule
        \multirow{3}{*}{\textbf{Paradigms}} &
        \multirow{3}{*}{\textbf{Models}} &
        \multicolumn{5}{c}{\textbf{Intra}} & \multicolumn{5}{c}{\textbf{Inter}}\\
        \cmidrule(lr){3-7} \cmidrule(lr){8-12}
        & & \multicolumn{2}{c}{\textbf{1$\sim$2-shot}} & \multicolumn{2}{c}{\textbf{5$\sim$10-shot}} & \multirow{2}{*}{\textbf{Avg.}} & \multicolumn{2}{c}{\textbf{1$\sim$2-shot}} & \multicolumn{2}{c}{\textbf{5$\sim$10-shot}} & \multirow{2}{*}{\textbf{Avg.}}\\

        \cmidrule(lr){3-4}\cmidrule(lr){5-6} \cmidrule(lr){8-9}\cmidrule(lr){10-11}

         & & 5 way & 10 way & 5 way & 10 way & & 5 way & 10 way & 5 way & 10 way & \\

         \cmidrule(lr){1-2}\cmidrule(lr){3-7} \cmidrule(lr){8-12}

        \multirow{5}{*}{\emph{One-stage}} &
         ProtoBERT & 20.76{\small{\textpm0.84}} & 15.05{\small{\textpm0.44}} & 42.54{\small{\textpm0.94}} & 35.40{\small{\textpm0.13}} & 28.44
         & 38.83{\small{\textpm1.49}} & 32.45{\small{\textpm0.79}} & 58.79{\small{\textpm0.44}} & 52.92{\small{\textpm0.37}} & 45.75\\

         & NNShot & 25.78{\small{\textpm0.91}} & 18.27{\small{\textpm0.41}} & 36.18{\small{\textpm0.79}} & 27.38{\small{\textpm0.53}} & 26.90
         & 47.24{\small{\textpm1.00}} & 38.87{\small{\textpm0.21}} & 55.64{\small{\textpm0.63}} & 49.57{\small{\textpm2.73}} & 47.83\\

         & StructShot & 30.21{\small{\textpm0.90}} & 	21.03{\small{\textpm1.13}} & 38.00{\small{\textpm1.29}} & 26.42{\small{\textpm0.60}} & 28.92
         & 51.88{\small{\textpm0.69}} & 43.34{\small{\textpm0.10}} & 57.32{\small{\textpm0.63}} & 49.57{\small{\textpm3.08}} & 50.53\\

        & FSLS & 30.38{\small{\textpm2.85}} & 28.31{\small{\textpm4.03}} & 46.85{\small{\textpm3.49}} & 40.76{\small{\textpm3.18}} & 36.58  
         & 44.52{\small{\textpm4.59}} & 44.01{\small{\textpm3.35}} & 59.74{\small{\textpm2.51}} & 56.67{\small{\textpm1.75}} &  51.24\\

        & CONTaiNER & 41.51{\small{\textpm0.07}} & 	36.62{\small{\textpm0.04}} & 57.83{\small{\textpm0.01}} & 51.04{\small{\textpm0.24}} & 46.75
         & 50.92{\small{\textpm0.29}} & 47.02{\small{\textpm0.24}} & 63.35{\small{\textpm0.07}} & 60.14{\small{\textpm0.16}} & 55.36\\	

         \cmidrule(lr){1-2}\cmidrule(lr){3-7} \cmidrule(lr){8-12}
         \multirow{6}{*}{\emph{Two-stage}} &

        ESD & 36.08{\small{\textpm1.60}} & 	30.00{\small{\textpm0.70}} & 52.14{\small{\textpm1.50}} & 42.15{\small{\textpm2.60}} & 40.09
         & 59.29{\small{\textpm1.25}} & 52.16{\small{\textpm0.79}} & {69.06{\small{\textpm0.80}}} & 	64.00{\small{\textpm0.43}} & 61.13\\

        & DecomposedMetaNER & 49.48{\small{\textpm0.85}} & 42.84{\small{\textpm0.46}} & 62.92{\small{\textpm0.57}} & 57.31{\small{\textpm0.25}} & 53.14
         &64.75{\small{\textpm0.35}} & 58.65{\small{\textpm0.43}} & 71.49{\small{\textpm0.47}} & 68.11{\small{\textpm0.05}} & 65.75\\

        & TadNER & \underline{60.78{\small{\textpm}0.32}} & \underline{55.44{\small{\textpm}0.08}} & \underline{67.94{\small{\textpm}0.17}} & \underline{60.87{\small{\textpm}0.22}} & \underline{61.26}
         & 64.83{\small{\textpm}0.14} & 64.06{\small{\textpm}0.19} & 72.12{\small{\textpm}0.12} &  \underline{69.94{\small{\textpm}0.15}} & 67.74\\

        & TSFNER & 56.35{\small{\textpm}0.64} & 50.51{\small{\textpm}0.36} & 65.22{\small{\textpm}0.52} & 58.35{\small{\textpm}0.19} & 57.61
         & 68.20{\small{\textpm}0.79} & 64.72{\small{\textpm}0.23} &  \underline{72.86{\small{\textpm}0.46}} & 68.62{\small{\textpm}0.27} & 68.60\\
         
        & BDCP & {53.96\small{\textpm}0.92} & {52.17\small{\textpm}0.56} & {59.25\small{\textpm}0.28} & {56.91\small{\textpm}1.12} & 55.57 
         &  \textbf{{69.68\small{\textpm}1.50}} &  \underline{{67.15\small{\textpm}0.28}} & {71.12\small{\textpm}0.97} & {68.13\small{\textpm}0.55} & \underline{69.02}\\  
         
        & \textbf{\ours} & \textbf{64.95{\small{\textpm}0.85}} & \textbf{61.24{\small{\textpm}0.82}} & \textbf{72.14{\small{\textpm}0.33}} & \textbf{67.53{\small{\textpm}0.12}} & \textbf{66.47}
         &  \underline{69.26{\small{\textpm}0.94}} & \textbf{67.43{\small{\textpm}0.35}} & \textbf{76.53{\small{\textpm}0.51}} & \textbf{72.24{\small{\textpm}0.22}} & \textbf{71.37}\\
         
        \bottomrule
    \end{tabular}
    }
    \caption{F1 scores with standard deviations on Few-NERD. The best results are in \textbf{bold} and the second best ones are \underline{underlined}. 
    }
    \label{tab:performance_comparison_fewnerd}
\end{table*}

\begin{table*}[ht]
    \centering
    \setlength{\tabcolsep}{2mm}
    \resizebox{\linewidth}{!}{
    \begin{tabular}{ll  ccccc  ccccc}
    \toprule
        \multirow{2}{*}{\textbf{Paradigms}} &
        \multirow{2}{*}{\textbf{Models}} &
        \multicolumn{5}{c}{\textbf{1-shot}} & \multicolumn{5}{c}{\textbf{5-shot}}\\
        \cmidrule(lr){3-7} \cmidrule(lr){8-12}

        & & \textbf{I2B2} & \textbf{CoNLL} & \textbf{WNUT} & \textbf{GUM} & \textbf{Avg.} & \textbf{I2B2} & \textbf{CoNLL} & \textbf{WNUT} & \textbf{GUM} & \textbf{Avg.} \\

        \cmidrule(lr){1-2}\cmidrule(lr){3-7} \cmidrule(lr){8-12}

        \multirow{6}{*}{\emph{One-stage}} &
        ProtoBERT & 13.4\small{\textpm 3.0} & 49.9\small{\textpm 8.6} & 17.4\small{\textpm 4.9} & 17.8\small{\textpm 3.5} & 24.6 
        & 17.9\small{\textpm 1.8} & 61.3\small{\textpm 9.1} & 22.8\small{\textpm 4.5} & 19.5\small{\textpm 3.4} & 30.4\\

        & NNShot & 15.3\small{\textpm 1.6} & 61.2\small{\textpm 10.4} & 22.7\small{\textpm 7.4} & 10.5\small{\textpm 2.9} & 27.4 
        & 22.0\small{\textpm 1.5} & 74.1\small{\textpm 2.3} & 27.3\small{\textpm 5.4} & 15.9\small{\textpm 1.8} & 34.8\\

        & StructShot & 21.4\small{\textpm 3.8} & 62.4\small{\textpm 10.5} & 24.2\small{\textpm 8.0} & 7.8\small{\textpm 2.1} & 29.0  
        & 30.3\small{\textpm 2.1} & 74.8\small{\textpm 2.4} & 30.4\small{\textpm 6.5} & 13.3\small{\textpm 1.3} & 37.2 \\

        & FSLS & 18.3\small{\textpm 3.5} & {50.9\small{\textpm 6.5}} & 14.3\small{\textpm 5.5} & 12.6\small{\textpm 2.8} &  24.0
        & 25.4\small{\textpm 2.7} & 63.9\small{\textpm 3.3} & 24.0\small{\textpm 3.2} & 18.8	\small{\textpm 2.2} &   33.1\\

        & CONTaiNER & 21.5\small{\textpm 1.7} & {61.2\small{\textpm 10.7}} & {27.5\small{\textpm 1.9}} & {18.5\small{\textpm 4.9}} & 32.2
        & 36.7\small{\textpm 2.1} & 75.8\small{\textpm 2.7} & 32.5\small{\textpm 3.8} & {25.2\small{\textpm 2.7}} & 42.6\\

        & MANNER & 24.3\small{\textpm 2.1} & 48.8\small{\textpm 3.5} & 27.9\small{\textpm 1.8} & 23.1\small{\textpm 2.3} & 31.0  
        & 33.9\small{\textpm }2.0 & 68.7\small{\textpm }3.2 & \underline{34.9\small{\textpm }} 2.5& \underline{40.7\small{\textpm }1.2} &  44.6 \\

        \cmidrule(lr){1-2}\cmidrule(lr){3-7} \cmidrule(lr){8-12}

        \multirow{5}{*}{\emph{Two-stage}} &
        DecomposedMetaNER & {15.5\small{\textpm 3.0}} & {61.2\small{\textpm 9.2}} & 27.7\small{\textpm 5.3} & 20.3\small{\textpm 4.2} & 31.2
        & 19.8\small{\textpm 2.6} & {75.2\small{\textpm 5.8}} & { 29.8\small{\textpm 3.9 }} & 33.5\small{\textpm 2.4} & 39.6\\

        & TadNER & \underline{39.3\small{\textpm 3.8}} & \underline{70.4\small{\textpm 10.6}}  & \underline{32.8\small{\textpm 4.8}} & 24.2\small{\textpm 4.1} & \underline{41.7}
        & \underline{45.2\small{\textpm 2.3}} & \underline{80.5\small{\textpm 3.6}} & 34.5\small{\textpm 4.6}  & 35.1\small{\textpm 2.2} & \underline{48.8}\\
         
        & TSFNER & {35.0\small{\textpm}0.9} & {62.5\small{\textpm}4.1} & {28.3\small{\textpm}2.5} & \textbf{{32.3\small{\textpm}3.0}} & 39.5
         & {40.6\small{\textpm}2.5} & {72.4\small{\textpm}5.6} & {34.7\small{\textpm}2.4} &{38.9\small{\textpm}0.9} & 46.7\\
         
        & BDCP & {33.2\small{\textpm}3.1} & {63.9\small{\textpm}8.3} & {30.3\small{\textpm}2.0} & {31.1\small{\textpm}1.5} & 39.6
         & {37.7\small{\textpm}2.2} & {69.8\small{\textpm}8.9} & {34.0\small{\textpm}1.6} & {34.6\small{\textpm}1.5} & 44.0\\
         
        & \textbf{\ours} & \textbf{40.2\small{\textpm }1.0} & \textbf{72.6\small{\textpm }3.1}  & \textbf{34.1\small{\textpm }2.1} & \underline{29.3\small{\textpm }2.8} & \textbf{44.0}
        & \textbf{47.1\small{\textpm }2.2} & \textbf{81.2\small{\textpm }2.9} & \textbf{43.2\small{\textpm }1.2}  & \textbf{44.0\small{\textpm }0.9} & \textbf{53.9}\\
        \bottomrule
    \end{tabular}
    }
    \caption{F1 scores with standard deviations for Cross-Dataset.}
    \label{tab:domaintransfer}
    \vspace{-0mm}
\end{table*}

\subsection{Main Results}
Tables \ref{tab:performance_comparison_fewnerd} and \ref{tab:domaintransfer} present the comparative results between our method and baselines on the \textbf{Few-NERD} and \textbf{Cross-Dataset} benchmarks, respectively. Several key observations emerge:

1) Overall, two-stage methods consistently outperform one-stage methods, underscoring the efficacy of task decomposition in few-shot NER tasks.

2) \ours consistently outperforms all baselines in all settings, often exceeding the performance of the second-best models by a notable margin. In particular, in the challenging \textbf{intra} task, \ours achieves an average increase in the F1 score of 5. 2\%.

3) Furthermore, in the 1-shot and 5-shot \textbf{Cross-Dataset} settings, \ours outperforms baselines by 2.3\% and 5.1\%, respectively. These results underscore the robustness of \ours in addressing cross-domain few-shot NER challenges.

4) TadNER, a competitive model, exhibits significantly degraded performance under certain settings, such as GUM. This issue primarily arises from dense entity sentences where boundary perception between different entities becomes challenging. In contrast, \ours effectively mitigates this challenge through the boundary-aware contrastive learning strategy, enabling accurate detection of entity spans and achieving superior performance.\looseness-1

\subsection{Ablation Study}
To validate the effectiveness of the main components in \ours, we introduce the following variant baselines for the ablation study:

\ours \textit{w/o} Boundary-Aware Span Detection (BASD): This variant removes the boundary-aware contrastive learning at the span detection stage and directly extracts entity spans using LLMs.

\ours \textit{w/o} Domain-Adaptation LoRAHub (DAL): This variant removes the composition of different LoRA modules at the type classification stage, using a single LoRA module to classify entities instead.

\ours \textit{w/o} Span Detection Fine-Tuning (SDF): This variant skips the fine-tuning on the support set of the target domain at the span detection stage.

\ours \textit{w/o} Type Classification Fine-Tuning (TCF): This variant skips the fine-tuning on the support set of the target domain at the type classification stage.

\ours \textit{w/o ALL}: This variant performs the few-shot NER task using the original LLMs (e.g., LlaMA-2-7B) without any of the enhancements provided by \ours.


\begin{table}[htb]
    \centering
    \setlength{\tabcolsep}{0.8mm}
    \resizebox{\linewidth}{!}{
    \begin{tabular}{lccccccccc}
    \toprule
        \multirow{2}{*}{\textbf{Models}}  & \multicolumn{4}{c}{\textbf{1-shot}} & \multicolumn{4}{c}{\textbf{5-shot}} & \multirow{2}{*}{\textbf{Avg.}}\\
        \cmidrule(lr){2-5} \cmidrule(lr){6-9}
        & {\small{I2B2}} & {\small{CoNLL}}  & {\small{WNUT}} & {\small{GUM}} & {\small{I2B2}} & {\small{CoNLL}}  & {\small{WNUT}} & {\small{GUM}}\\
        \cmidrule(lr){1-1}   \cmidrule(lr){2-5} \cmidrule(lr){6-9}\cmidrule(lr){10-10}
        \textbf{\ours} & \textbf{40.2} & \textbf{72.6} & \textbf{34.1} & \textbf{29.3} & \textbf{47.1} & \textbf{81.2} & \textbf{43.2} & \textbf{44.0} & \textbf{49.0} \\
        \cmidrule(lr){1-1}    \cmidrule(lr){2-5} \cmidrule(lr){6-9}\cmidrule(lr){10-10}
        \textit{w/o} BASD & 22.7 & 65.7 & 30.7 & 26.1 & 30.1 & 73.9 & 39.0 & 39.3 & 40.9 \\
        \textit{w/o} DAL & 30.3 & 64.0 & 32.5 & 27.0 & 34.6 & 73.8 & 39.5 & 40.2 & 42.2 \\
        \textit{w/o} SDF & 37.3 & 68.8 & 31.2 & 28.1 & 45.0 & 76.5 & 40.3 & 42.2 & 46.2\\
        \textit{w/o} TCF & 39.2 & 69.0 & 32.1 & 28.0 & 45.7 & 78.2 & 40.9 & 42.4 & 46.9 \\
        \textit{w/o ALL} & 20.9 & 41.3 & 17.0 & 15.6 & 24.5 & 56.1 & 20.3 & 18.2 & 26.7 \\
    \bottomrule
    \end{tabular}
    }
    \caption{Ablation study results for    Cross-Dataset.}
    \label{tab:ablation}
    \vspace{-0mm}
\end{table}
From Table \ref{tab:ablation}, we observe the following:

1) The removal of the boundary-aware contrastive learning strategy results in a performance decline across most cases, particularly in entity-sparse datasets like I2B2, where many spans are falsely detected.

2) Omitting the domain-aware LoRAHub leads to a significant performance decrease. This indicates that our model effectively aligns a better prototype space for entity types, which is crucial in cross-domain scenarios.

3) Eliminating fine-tuning in both the span detection and type classification stages causes a minor performance drop. This demonstrates that the prototype in the source domain aligns well with the target domain, and that LLMs already possess good boundary perception abilities despite encountering different entity types in the target domain after training in the source domain.

4) Although LLMs exhibit superior performance in few-shot tasks compared to most pretrained models, they still lag behind our approach. The significant disparity compared to the original LlaMA-2-7B underscores our model's effective utilization of provided support samples from the target domain, thereby enhancing the performance of LLMs in few-shot scenarios.
\subsection{Examination of other LLMs}
To evaluate the generalizability of our enhanced entity boundary perception, we extend \ours to other mainstream open-source LLMs under the GUM 5-shot setting, including Mistral-7B \cite{jiang2023mistral} and LlaMA-3-8B.
\begin{figure}[htbp]
\vspace{-0mm}
\centering
\includegraphics[width=1.0\linewidth]{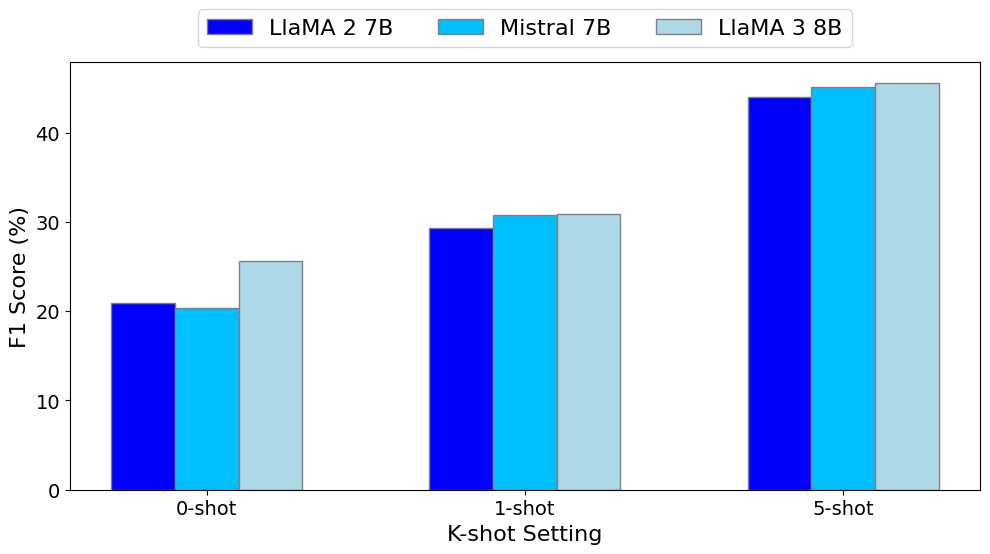}
\caption{F1 Score for different LLMs under the GUM 5-shot setting.}
\label{diff_models}
\vspace{-0mm}
\end{figure}
As shown in Figure \ref{diff_models}, substituting the LLM in \ours with these models leads to significant improvements in F1 scores for both 1-shot and 5-shot scenarios compared to the 0-shot baseline. This demonstrates the broad applicability and effectiveness of our method across different LLM architectures. 
\subsection{Impact of Different Hidden Layers}\label{sec:impact_layers}
\begin{figure}[htbp]
\vspace{-0mm}
\centering
\includegraphics[width=1\linewidth]{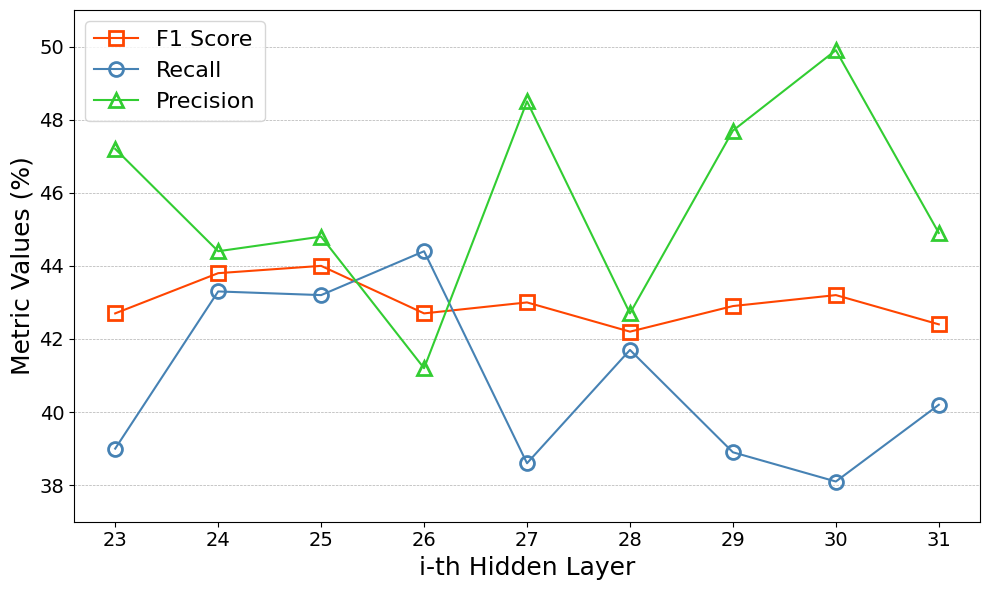}
\caption{F1 Score, Recall, and Precision for different hidden layers under the GUM 5-shot setting.}
\label{compare_layer}
\vspace{-0mm}
\end{figure}

To determine which hidden layer's output in LlaMA-2 captures higher-level abstract information for constructing a better boundary-aware feature space, we compare overall performance by calculating the contrastive learning loss across different hidden layers under the GUM 5-shot setting. The performance of different hidden layers is shown in Figure \ref{compare_layer}. We observe that the highest F1 score is achieved when calculating the contrastive learning loss on the 25th layer. Notably, unlike other layers where there is a significant disparity between recall and precision, the 25th layer exhibits a relatively small difference between these metrics.

\section{Conclusion}
In this paper, we propose the \ours framework for few-shot named entity recognition (NER), addressing entity span detection and entity type classification in two stages. For entity span detection, we introduce a boundary-aware contrastive learning strategy to minimize the distance between span embeddings of entities and their corresponding types using LLMs. Building on this, we employ domain adaptation with LoRAHub to construct more accurate prototypes that preserve and align entity type knowledge from the source domain during the entity classification stage. Extensive experiments demonstrate that \ours outperforms previous state-of-the-art methods and is applicable to various LLMs.


\section*{Limitations}
Our work has two main limitations: 1) \ours employs a single, specific prompt template for each stage, utilizing descriptive task instructions and limited answer options. However, there exist numerous alternative templates for generative language models. This limitation suggests the potential for future research to explore various prompt templates to enhance entity boundary detection and entity type understanding. 2) Limited access to high-performance computing facilities has prevented us from evaluating our approach on large LLMs, such as LlaMA-3-70B. This limitation highlights the potential for future work to investigate different model architectures for improved few-shot NER performance.


\bibliography{custom,few-shot-NER,dataset}

\begin{thebibliography}{48}
\expandafter\ifx\csname natexlab\endcsname\relax\def\natexlab#1{#1}\fi

\bibitem[{Buonocore et~al.(2023)Buonocore, Crema, Redolfi, Bellazzi, and Parimbelli}]{buonocore2023localizing}
Tommaso~Mario Buonocore, Claudio Crema, Alberto Redolfi, Riccardo Bellazzi, and Enea Parimbelli. 2023.
\newblock Localizing in-domain adaptation of transformer-based biomedical language models.
\newblock \emph{Journal of Biomedical Informatics}, 144:104431.

\bibitem[{Chiu and Nichols(2016)}]{chiu2016named}
Jason~P.C. Chiu and Eric Nichols. 2016.
\newblock \href {https://doi.org/10.1162/tacl_a_00104} {Named entity recognition with bidirectional {LSTM}-{CNN}s}.
\newblock \emph{Transactions of the Association for Computational Linguistics}, 4:357--370.

\bibitem[{Cui et~al.(2021)Cui, Wu, Liu, Yang, and Zhang}]{cui-etal-2021-template}
Leyang Cui, Yu~Wu, Jian Liu, Sen Yang, and Yue Zhang. 2021.
\newblock \href {https://doi.org/10.18653/v1/2021.findings-acl.161} {Template-based named entity recognition using {BART}}.
\newblock In \emph{Findings of the Association for Computational Linguistics: ACL-IJCNLP 2021}, pages 1835--1845, Online. Association for Computational Linguistics.

\bibitem[{Das et~al.(2022)Das, Katiyar, Passonneau, and Zhang}]{das-etal-2022-container}
Sarkar Snigdha~Sarathi Das, Arzoo Katiyar, Rebecca Passonneau, and Rui Zhang. 2022.
\newblock \href {https://doi.org/10.18653/v1/2022.acl-long.439} {{CONT}ai{NER}: Few-shot named entity recognition via contrastive learning}.
\newblock In \emph{Proceedings of the 60th Annual Meeting of the Association for Computational Linguistics (Volume 1: Long Papers)}, pages 6338--6353, Dublin, Ireland. Association for Computational Linguistics.

\bibitem[{Derczynski et~al.(2017)Derczynski, Nichols, van Erp, and Limsopatham}]{derczynski2017wnut17}
Leon Derczynski, Eric Nichols, Marieke van Erp, and Nut Limsopatham. 2017.
\newblock \href {https://doi.org/10.18653/v1/W17-4418} {Results of the {WNUT}2017 shared task on novel and emerging entity recognition}.
\newblock In \emph{Proceedings of the 3rd Workshop on Noisy User-generated Text}, pages 140--147, Copenhagen, Denmark. Association for Computational Linguistics.

\bibitem[{Ding et~al.(2021)Ding, Xu, Chen, Wang, Han, Xie, Zheng, and Liu}]{ding-etal-2021-nerd}
Ning Ding, Guangwei Xu, Yulin Chen, Xiaobin Wang, Xu~Han, Pengjun Xie, Haitao Zheng, and Zhiyuan Liu. 2021.
\newblock \href {https://doi.org/10.18653/v1/2021.acl-long.248} {Few-{NERD}: A few-shot named entity recognition dataset}.
\newblock In \emph{Proceedings of the 59th Annual Meeting of the Association for Computational Linguistics and the 11th International Joint Conference on Natural Language Processing (Volume 1: Long Papers)}, pages 3198--3213, Online. Association for Computational Linguistics.

\bibitem[{Fang et~al.(2023)Fang, Wang, Meng, Xie, Huang, and Jiang}]{fang2023manner}
Jinyuan Fang, Xiaobin Wang, Zaiqiao Meng, Pengjun Xie, Fei Huang, and Yong Jiang. 2023.
\newblock Manner: A variational memory-augmented model for cross domain few-shot named entity recognition.
\newblock In \emph{Proceedings of the 61st Annual Meeting of the Association for Computational Linguistics (Volume 1: Long Papers)}, pages 4261--4276.

\bibitem[{Fritzler et~al.(2019)Fritzler, Logacheva, and Kretov}]{fritzler-2019}
Alexander Fritzler, Varvara Logacheva, and Maksim Kretov. 2019.
\newblock \href {https://doi.org/10.1145/3297280.3297378} {Few-shot classification in named entity recognition task}.
\newblock In \emph{Proceedings of the 34th ACM/SIGAPP Symposium on Applied Computing}, SAC '19, page 993–1000, New York, NY, USA. Association for Computing Machinery.

\bibitem[{Grangier and Iter(2022)}]{grangier2022trade}
David Grangier and Dan Iter. 2022.
\newblock The trade-offs of domain adaptation for neural language models.
\newblock In \emph{Proceedings of the 60th Annual Meeting of the Association for Computational Linguistics (Volume 1: Long Papers)}, pages 3802--3813.

\bibitem[{Guo et~al.(2021)Guo, Rennard, Xypolopoulos, and Vazirgiannis}]{guo2021bertweetfr}
Yanzhu Guo, Virgile Rennard, Christos Xypolopoulos, and Michalis Vazirgiannis. 2021.
\newblock Bertweetfr: Domain adaptation of pre-trained language models for french tweets.
\newblock In \emph{Proceedings of the Seventh Workshop on Noisy User-generated Text (W-NUT 2021)}, pages 445--450. Association for Computational Linguistics.

\bibitem[{Huang et~al.(2023)Huang, Liu, Lin, Pang, Du, and Lin}]{huang2023lorahub}
Chengsong Huang, Qian Liu, Bill~Yuchen Lin, Tianyu Pang, Chao Du, and Min Lin. 2023.
\newblock Lorahub: Efficient cross-task generalization via dynamic lora composition.
\newblock \emph{arXiv preprint arXiv:2307.13269}.

\bibitem[{Huang et~al.(2022)Huang, He, Wang, Zhang, Gong, Mao, and Li}]{huang-etal-2022-copner}
Yucheng Huang, Kai He, Yige Wang, Xianli Zhang, Tieliang Gong, Rui Mao, and Chen Li. 2022.
\newblock \href {https://aclanthology.org/2022.coling-1.222} {{COPNER}: Contrastive learning with prompt guiding for few-shot named entity recognition}.
\newblock In \emph{Proceedings of the 29th International Conference on Computational Linguistics}, pages 2515--2527, Gyeongju, Republic of Korea. International Committee on Computational Linguistics.

\bibitem[{Huang et~al.(2015)Huang, Xu, and Yu}]{huang2015bidirectional}
Zhiheng Huang, Wei Xu, and Kai Yu. 2015.
\newblock \href {https://arxiv.org/abs/1508.01991} {Bidirectional lstm-crf models for sequence tagging}.
\newblock \emph{arXiv preprint arXiv:1508.01991}.

\bibitem[{Ji et~al.(2022)Ji, Li, Gan, Yu, Ma, Liu, and Yang}]{ji-etal-2022-shot}
Bin Ji, Shasha Li, Shaoduo Gan, Jie Yu, Jun Ma, Huijun Liu, and Jing Yang. 2022.
\newblock \href {https://aclanthology.org/2022.coling-1.159} {Few-shot named entity recognition with entity-level prototypical network enhanced by dispersedly distributed prototypes}.
\newblock In \emph{Proceedings of the 29th International Conference on Computational Linguistics}, pages 1842--1854, Gyeongju, Republic of Korea. International Committee on Computational Linguistics.

\bibitem[{Ji and Kong(2024)}]{ji-kong-2024-novel-three}
Shengjie Ji and Fang Kong. 2024.
\newblock \href {https://aclanthology.org/2024.lrec-main.116} {A novel three-stage framework for few-shot named entity recognition}.
\newblock In \emph{Proceedings of the 2024 Joint International Conference on Computational Linguistics, Language Resources and Evaluation (LREC-COLING 2024)}, pages 1293--1305, Torino, Italia. ELRA and ICCL.

\bibitem[{Jiang et~al.(2023)Jiang, Sablayrolles, Mensch, Bamford, Chaplot, Casas, Bressand, Lengyel, Lample, Saulnier et~al.}]{jiang2023mistral}
Albert~Q Jiang, Alexandre Sablayrolles, Arthur Mensch, Chris Bamford, Devendra~Singh Chaplot, Diego de~las Casas, Florian Bressand, Gianna Lengyel, Guillaume Lample, Lucile Saulnier, et~al. 2023.
\newblock Mistral 7b.
\newblock \emph{arXiv preprint arXiv:2310.06825}.

\bibitem[{Khosla et~al.(2020)Khosla, Teterwak, Wang, Sarna, Tian, Isola, Maschinot, Liu, and Krishnan}]{NEURIPS2020_d89a66c7_supervised_contrastive_learning}
Prannay Khosla, Piotr Teterwak, Chen Wang, Aaron Sarna, Yonglong Tian, Phillip Isola, Aaron Maschinot, Ce~Liu, and Dilip Krishnan. 2020.
\newblock \href {https://proceedings.neurips.cc/paper/2020/file/d89a66c7c80a29b1bdbab0f2a1a94af8-Paper.pdf} {Supervised contrastive learning}.
\newblock In \emph{Advances in Neural Information Processing Systems}, volume~33, pages 18661--18673. Curran Associates, Inc.

\bibitem[{Lample et~al.(2016)Lample, Ballesteros, Subramanian, Kawakami, and Dyer}]{lample2016neural}
Guillaume Lample, Miguel Ballesteros, Sandeep Subramanian, Kazuya Kawakami, and Chris Dyer. 2016.
\newblock \href {https://doi.org/10.18653/v1/N16-1030} {Neural architectures for named entity recognition}.
\newblock In \emph{Proceedings of the 2016 Conference of the North {A}merican Chapter of the Association for Computational Linguistics: Human Language Technologies}, pages 260--270, San Diego, California. Association for Computational Linguistics.

\bibitem[{Lee et~al.(2022)Lee, Kadakia, Tan, Agarwal, Feng, Shibuya, Mitani, Sekiya, Pujara, and Ren}]{lee-etal-2022-good}
Dong-Ho Lee, Akshen Kadakia, Kangmin Tan, Mahak Agarwal, Xinyu Feng, Takashi Shibuya, Ryosuke Mitani, Toshiyuki Sekiya, Jay Pujara, and Xiang Ren. 2022.
\newblock \href {https://doi.org/10.18653/v1/2022.acl-long.192} {Good examples make a faster learner: Simple demonstration-based learning for low-resource {NER}}.
\newblock In \emph{Proceedings of the 60th Annual Meeting of the Association for Computational Linguistics (Volume 1: Long Papers)}, pages 2687--2700, Dublin, Ireland. Association for Computational Linguistics.

\bibitem[{Li et~al.(2023)Li, Yu, and Qian}]{li-etal-2023-type-aware}
Yongqi Li, Yu~Yu, and Tieyun Qian. 2023.
\newblock \href {https://doi.org/10.18653/v1/2023.findings-emnlp.598} {Type-aware decomposed framework for few-shot named entity recognition}.
\newblock In \emph{Findings of the Association for Computational Linguistics: EMNLP 2023}, pages 8911--8927, Singapore. Association for Computational Linguistics.

\bibitem[{Loshchilov and Hutter(2019)}]{loshchilov2018decoupled}
Ilya Loshchilov and Frank Hutter. 2019.
\newblock \href {https://openreview.net/forum?id=Bkg6RiCqY7} {Decoupled weight decay regularization}.
\newblock In \emph{7th International Conference on Learning Representations, {ICLR} 2019, New Orleans, LA, USA, May 6-9, 2019}. OpenReview.net.

\bibitem[{Ma et~al.(2022{\natexlab{a}})Ma, Zhou, Gui, Tan, Li, Zhang, and Huang}]{ma-etal-2022-template}
Ruotian Ma, Xin Zhou, Tao Gui, Yiding Tan, Linyang Li, Qi~Zhang, and Xuanjing Huang. 2022{\natexlab{a}}.
\newblock \href {https://doi.org/10.18653/v1/2022.naacl-main.420} {Template-free prompt tuning for few-shot {NER}}.
\newblock In \emph{Proceedings of the 2022 Conference of the North American Chapter of the Association for Computational Linguistics: Human Language Technologies}, pages 5721--5732, Seattle, United States. Association for Computational Linguistics.

\bibitem[{Ma et~al.(2022{\natexlab{b}})Ma, Jiang, Wu, Zhao, and Lin}]{ma-etal-2022-decomposed}
Tingting Ma, Huiqiang Jiang, Qianhui Wu, Tiejun Zhao, and Chin-Yew Lin. 2022{\natexlab{b}}.
\newblock \href {https://doi.org/10.18653/v1/2022.findings-acl.124} {Decomposed meta-learning for few-shot named entity recognition}.
\newblock In \emph{Findings of the Association for Computational Linguistics: ACL 2022}, pages 1584--1596, Dublin, Ireland. Association for Computational Linguistics.

\bibitem[{Ma and Hovy(2016)}]{ma2016end}
Xuezhe Ma and Eduard Hovy. 2016.
\newblock \href {https://doi.org/10.18653/v1/P16-1101} {End-to-end sequence labeling via bi-directional {LSTM}-{CNN}s-{CRF}}.
\newblock In \emph{Proceedings of the 54th Annual Meeting of the Association for Computational Linguistics (Volume 1: Long Papers)}, pages 1064--1074, Berlin, Germany. Association for Computational Linguistics.

\bibitem[{Ma et~al.(2023)Ma, Cao, Hong, and Sun}]{ma2023large}
Yubo Ma, Yixin Cao, Yong Hong, and Aixin Sun. 2023.
\newblock Large language model is not a good few-shot information extractor, but a good reranker for hard samples!
\newblock In \emph{Findings of the Association for Computational Linguistics: EMNLP 2023}, pages 10572--10601.

\bibitem[{Rietzler et~al.(2020)Rietzler, Stabinger, Opitz, and Engl}]{rietzler2020adapt}
Alexander Rietzler, Sebastian Stabinger, Paul Opitz, and Stefan Engl. 2020.
\newblock Adapt or get left behind: Domain adaptation through bert language model finetuning for aspect-target sentiment classification.
\newblock In \emph{Proceedings of the Twelfth Language Resources and Evaluation Conference}, pages 4933--4941.

\bibitem[{Ryu et~al.(2022)Ryu, Lee, and Lee}]{ryu2022knowledge}
Minho Ryu, Geonseok Lee, and Kichun Lee. 2022.
\newblock Knowledge distillation for bert unsupervised domain adaptation.
\newblock \emph{Knowledge and Information Systems}, 64(11):3113--3128.

\bibitem[{Sachidananda et~al.(2021)Sachidananda, Kessler, and Lai}]{sachidananda2021efficient}
Vin Sachidananda, Jason Kessler, and Yi-An Lai. 2021.
\newblock Efficient domain adaptation of language models via adaptive tokenization.
\newblock In \emph{Proceedings of the Second Workshop on Simple and Efficient Natural Language Processing}, pages 155--165.

\bibitem[{Shen et~al.(2021)Shen, Ma, Tan, Zhang, Wang, and Lu}]{shen-etal-2021-locate}
Yongliang Shen, Xinyin Ma, Zeqi Tan, Shuai Zhang, Wen Wang, and Weiming Lu. 2021.
\newblock \href {https://doi.org/10.18653/v1/2021.acl-long.216} {Locate and label: A two-stage identifier for nested named entity recognition}.
\newblock In \emph{Proceedings of the 59th Annual Meeting of the Association for Computational Linguistics and the 11th International Joint Conference on Natural Language Processing (Volume 1: Long Papers)}, pages 2782--2794, Online. Association for Computational Linguistics.

\bibitem[{Snell et~al.(2017)Snell, Swersky, and Zemel}]{snell_prototypical_2017}
Jake Snell, Kevin Swersky, and Richard Zemel. 2017.
\newblock \href {https://proceedings.neurips.cc/paper/2017/file/cb8da6767461f2812ae4290eac7cbc42-Paper.pdf} {Prototypical {Networks} for {Few}-shot {Learning}}.
\newblock In \emph{Advances in {Neural} {Information} {Processing} {Systems}}, volume~30. Curran Associates, Inc.

\bibitem[{Stubbs and Uzuner(2015)}]{stubbs-2015-i2b2-2014}
Amber Stubbs and {\"O}zlem Uzuner. 2015.
\newblock \href {https://www.sciencedirect.com/science/article/pii/S1532046415001823} {Annotating longitudinal clinical narratives for de-identification: The 2014 i2b2/uthealth corpus}.
\newblock \emph{Journal of biomedical informatics}, 58:S20--S29.

\bibitem[{Tjong Kim~Sang and De~Meulder(2003)}]{sang2003conll2003}
Erik~F. Tjong Kim~Sang and Fien De~Meulder. 2003.
\newblock \href {https://aclanthology.org/W03-0419} {Introduction to the {C}o{NLL}-2003 shared task: Language-independent named entity recognition}.
\newblock In \emph{Proceedings of the Seventh Conference on Natural Language Learning at {HLT}-{NAACL} 2003}, pages 142--147.

\bibitem[{Wang et~al.(2022)Wang, Xu, Liu, Zhou, Cao, Chang, and Sui}]{wang-2022-enhanced}
Peiyi Wang, Runxin Xu, Tianyu Liu, Qingyu Zhou, Yunbo Cao, Baobao Chang, and Zhifang Sui. 2022.
\newblock \href {https://doi.org/10.18653/v1/2022.naacl-main.369} {An enhanced span-based decomposition method for few-shot sequence labeling}.
\newblock In \emph{Proceedings of the 2022 Conference of the North American Chapter of the Association for Computational Linguistics: Human Language Technologies}, pages 5012--5024, Seattle, United States. Association for Computational Linguistics.

\bibitem[{Wang et~al.(2023)Wang, Sun, Li, Ouyang, Wu, Zhang, Li, and Wang}]{wang2023gpt}
Shuhe Wang, Xiaofei Sun, Xiaoya Li, Rongbin Ouyang, Fei Wu, Tianwei Zhang, Jiwei Li, and Guoyin Wang. 2023.
\newblock Gpt-ner: Named entity recognition via large language models.
\newblock \emph{arXiv preprint arXiv:2304.10428}.

\bibitem[{Wang et~al.(2021)Wang, Chu, Zhang, and Gao}]{wang2021learning-from-description}
Yaqing Wang, Haoda Chu, Chao Zhang, and Jing Gao. 2021.
\newblock \href {https://doi.org/10.18653/v1/2021.findings-emnlp.139} {Learning from language description: Low-shot named entity recognition via decomposed framework}.
\newblock In \emph{Findings of the Association for Computational Linguistics: EMNLP 2021}, pages 1618--1630, Punta Cana, Dominican Republic. Association for Computational Linguistics.

\bibitem[{Weischedel et~al.(2013)Weischedel, Palmer, Marcus, Hovy, Pradhan, Ramshaw, Xue, Taylor, Kaufman, Franchini et~al.}]{weischedel2013ontonotes}
Ralph Weischedel, Martha Palmer, Mitchell Marcus, Eduard Hovy, Sameer Pradhan, Lance Ramshaw, Nianwen Xue, Ann Taylor, Jeff Kaufman, Michelle Franchini, et~al. 2013.
\newblock \href {https://catalog.ldc.upenn.edu/LDC2013T19} {Ontonotes release 5.0 ldc2013t19}.
\newblock \emph{Linguistic Data Consortium, Philadelphia, PA}.

\bibitem[{Wiseman and Stratos(2019)}]{wiseman2019label}
Sam Wiseman and Karl Stratos. 2019.
\newblock Label-agnostic sequence labeling by copying nearest neighbors.
\newblock In \emph{Proceedings of the 57th Annual Meeting of the Association for Computational Linguistics}, pages 5363--5369.

\bibitem[{Xie et~al.(2023)Xie, Li, Zhang, Zhang, Liu, and Wang}]{xie2023empirical}
Tingyu Xie, Qi~Li, Jian Zhang, Yan Zhang, Zuozhu Liu, and Hongwei Wang. 2023.
\newblock Empirical study of zero-shot ner with chatgpt.
\newblock In \emph{Proceedings of the 2023 Conference on Empirical Methods in Natural Language Processing}, pages 7935--7956.

\bibitem[{Xue et~al.(2024)Xue, Zhang, Xu, and Niu}]{xue2024robust}
Xiaojun Xue, Chunxia Zhang, Tianxiang Xu, and Zhendong Niu. 2024.
\newblock Robust few-shot named entity recognition with boundary discrimination and correlation purification.
\newblock In \emph{Proceedings of the AAAI Conference on Artificial Intelligence}, volume~38, pages 19341--19349.

\bibitem[{Yang and Katiyar(2020)}]{yang-katiyar-2020-simple}
Yi~Yang and Arzoo Katiyar. 2020.
\newblock \href {https://doi.org/10.18653/v1/2020.emnlp-main.516} {Simple and effective few-shot named entity recognition with structured nearest neighbor learning}.
\newblock In \emph{Proceedings of the 2020 Conference on Empirical Methods in Natural Language Processing (EMNLP)}, pages 6365--6375, Online. Association for Computational Linguistics.

\bibitem[{Zeldes(2017)}]{zeldes2017gum}
Amir Zeldes. 2017.
\newblock \href {https://doi.org/10.1007/s10579-016-9343-x} {The gum corpus: Creating multilayer resources in the classroom}.
\newblock \emph{Lang. Resour. Eval.}, 51(3):581–612.

\bibitem[{Zhang et~al.(2022)Zhang, Yu, Wang, Liu, Su, and Xu}]{zhang2022exploring}
Xinghua Zhang, Bowen Yu, Yubin Wang, Tingwen Liu, Taoyu Su, and Hongbo Xu. 2022.
\newblock \href {https://dl.acm.org/doi/abs/10.1145/3477495.3531976} {Exploring modular task decomposition in cross-domain named entity recognition}.
\newblock In \emph{Proceedings of the 45th International ACM SIGIR Conference on Research and Development in Information Retrieval}, pages 301--311.

\bibitem[{Zhang et~al.(2021)Zhang, Zhang, Deng, Li, Jia, and Zhang}]{zhang2021semi}
Yabin Zhang, Haojian Zhang, Bin Deng, Shuai Li, Kui Jia, and Lei Zhang. 2021.
\newblock Semi-supervised models are strong unsupervised domain adaptation learners.
\newblock \emph{arXiv preprint arXiv:2106.00417}.

\bibitem[{Zhang et~al.(2024{\natexlab{a}})Zhang, Chen, Bai, Kang, Guo, and Zhang}]{zhang2024question}
Yu~Zhang, Kehai Chen, Xuefeng Bai, Zhao Kang, Quanjiang Guo, and Min Zhang. 2024{\natexlab{a}}.
\newblock Question-guided knowledge graph re-scoring and injection for knowledge graph question answering.
\newblock In \emph{Findings of the Association for Computational Linguistics: EMNLP 2024}, pages 8972--8985.

\bibitem[{Zhang and Kang(2024)}]{zhang2024tpn}
Yu~Zhang and Zhao Kang. 2024.
\newblock Tpn: Transferable proto-learning network towards few-shot document-level relation extraction.
\newblock In \emph{2024 International Joint Conference on Neural Networks (IJCNN)}, pages 1--9. IEEE.

\bibitem[{Zhang et~al.(2024{\natexlab{b}})Zhang, Zhao, Gao, and Hu}]{zhang2024linkner}
Zhen Zhang, Yuhua Zhao, Hang Gao, and Mengting Hu. 2024{\natexlab{b}}.
\newblock Linkner: Linking local named entity recognition models to large language models using uncertainty.
\newblock In \emph{Proceedings of the ACM on Web Conference 2024}, pages 4047--4058.

\bibitem[{Zhou et~al.(2024)Zhou, Zhang, Gu, Chen, and Poon}]{zhou2024universalner}
Wenxuan Zhou, Sheng Zhang, Yu~Gu, Muhao Chen, and Hoifung Poon. 2024.
\newblock \href {https://openreview.net/forum?id=r65xfUb76p} {Universal{NER}: Targeted distillation from large language models for open named entity recognition}.
\newblock In \emph{The Twelfth International Conference on Learning Representations}.

\bibitem[{Zhu et~al.(2024)Zhu, Kang, and Hui}]{zhu2024fcds}
Xudong Zhu, Zhao Kang, and Bei Hui. 2024.
\newblock Fcds: Fusing constituency and dependency syntax into document-level relation extraction.
\newblock In \emph{Proceedings of the 2024 Joint International Conference on Computational Linguistics, Language Resources and Evaluation (LREC-COLING 2024)}, pages 7141--7152.

\end{thebibliography}
\bibliographystyle{acl_natbib}

\clearpage

\appendix
\section{Appendix}\label{appendix_prompt}
\begin{figure*}[h]
\centering
\includegraphics[width=1\textwidth]{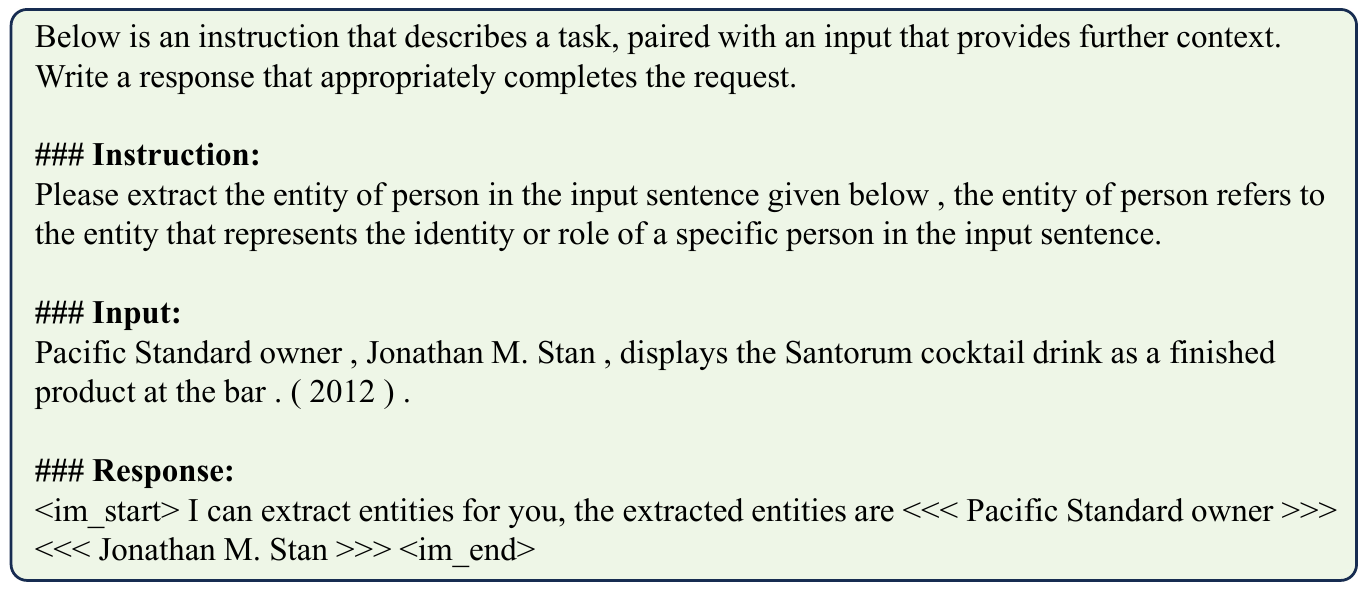}
\caption{Example of the prompt in entity span detection.}
\label{fig_prompt1}
\vspace{-3mm}
\end{figure*}

\begin{figure*}[h]
\centering
\includegraphics[width=1\textwidth]{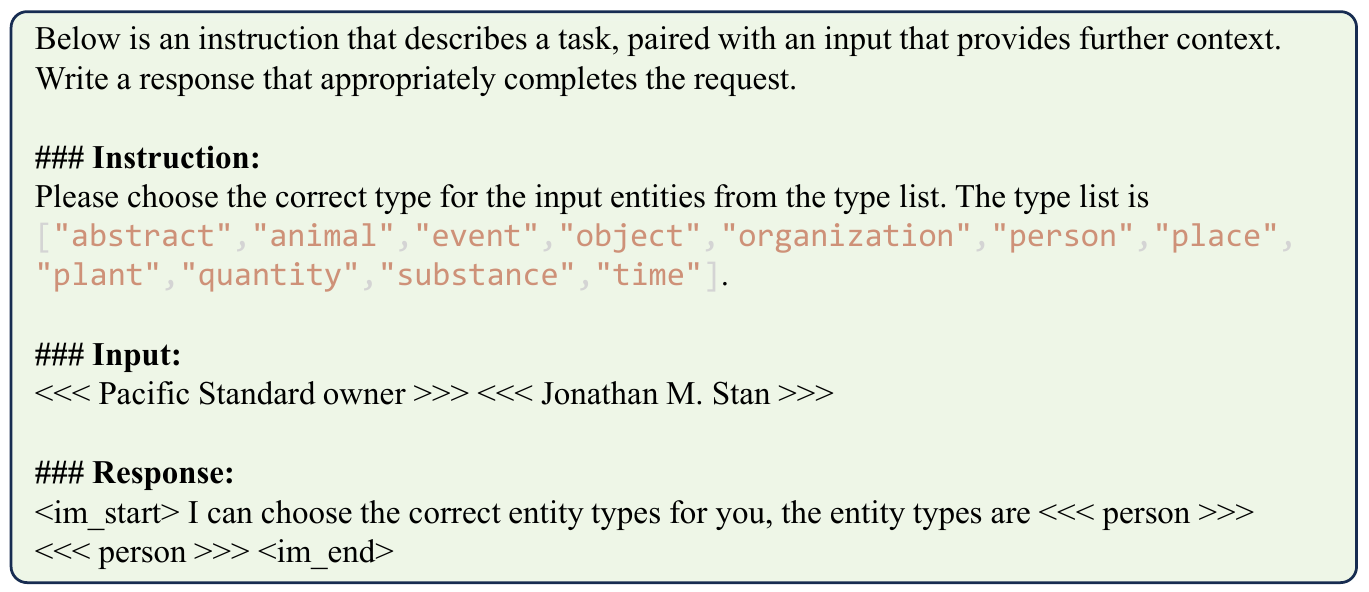}
\caption{Example of the prompt in entity entity classification.}
\label{fig_prompt2}
\vspace{-3mm}
\end{figure*}

\subsection{Examples of Prompt}\label{prompt}
Figures \ref{fig_prompt1} and \ref{fig_prompt2} provide examples of the prompts used in the two stages of our method. To tailor these prompts to our task, we design a specific output format for the LLM. Each output starts with $<$im\_start$>$ and ends with $<$im\_end$>$. For instances involving multiple entity spans and types, we encapsulate them together using $<<<$ $>>>$.

\subsection{Baselines}\label{appendix_baselines}

1) \emph{one-stage methods}:
\vspace{-3mm}
\begin{itemize}
    \item \textbf{ProtoBERT}~\cite{snell_prototypical_2017}is a popular few-shot method built on prototypical networks, utilizing BERT as its backbone;
    \vspace{-3mm}
    \item \textbf{NNShot}~\cite{wiseman2019label} is a straightforward approach that utilizes token-level nearest neighbor classification;
    \vspace{-3mm}
    \item \textbf{StructShot}~\cite{yang-katiyar-2020-simple} 
adopts an additional Viterbi decoder on top of NNShot;
    \vspace{-3mm}
    \item \textbf{CONTaiNER}~\cite{das-etal-2022-container} leverages contrastive learning to infer the distributional distance between Gaussian embeddings of entities;
    \vspace{-3mm}
    \item \textbf{MANNER}~\cite{fang2023manner} uses a memory module and optimal transport to adapt source domain information for few-shot tasks in the target domain.
\end{itemize}
\vspace{-3mm}
2) \emph{two-stage methods}:
\vspace{-3mm}
\begin{itemize}
    \item \textbf{ESD}~\cite{wang-2022-enhanced} enhances prototypical networks with inter- and cross-span attention , and introduces multiple prototypes for the O label;
    \vspace{-3mm}
    \item \textbf{DecomposedMetaNER}~\cite{ma-etal-2022-decomposed} integrates model-agnostic meta-learning into prototypical networks to more effectively leverage the support set;
    \vspace{-3mm}
    \item \textbf{TadNER}~\cite{li-etal-2023-type-aware} employs type-aware contrastive learning and span filtering to construct precise prototypes and eliminate false spans;
    \vspace{-3mm}
    \item \textbf{TSFNER}~\cite{ji-kong-2024-novel-three} incorporates a teacher span recognizer for generating soft labels, a student span recognizer, and a prompt-based entity classifier;
    \vspace{-3mm}
    \item \textbf{BDCP}~\cite{xue2024robust} introduces an entity boundary discriminative module for span detection and refines entity-context correlations to mitigate textual adversarial attacks.
\end{itemize}

\subsection{Detailed Type Names}\label{appendix_type_names}
Following \cite{li-etal-2023-type-aware}, we substitute the original dataset labels with their corresponding natural-language forms of type names employed in our prompt. Tables~\ref{tab:dataset_labels_nlf_1} and~\ref{tab:dataset_labels_nlf_2} present the detailed conversions for various datasets.

\clearpage

\begin{table}[htb]
\small
\begin{center}
\resizebox{\columnwidth}{!}  {
\begin{tabular}{lcl}
\toprule
\bf Dataset & \makecell[c]{\textbf{Labels}} & \makecell[c]{\textbf{Type names}} \\

\cmidrule(lr){1-1} \cmidrule(lr){2-3}
\multirow{66}{*}{\textbf{Few-NERD} } 
    &art-broadcastprogram &  broadcast program \\
    &art-film & film \\
    &art-music & music \\
    &art-other & other art\\
    &art-painting & painting \\
    &art-writtenart & written art \\
    &person-actor & actor \\
    &person-artist/author & artist author \\
    &person-athlete & athlete \\
    &person-director & director \\
    &person-other & other person \\
    &person-politician & politician \\
    &person-scholar & scholar \\
    &person-soldier &  soldier \\ 
    &product-airplane & airplane \\
    &product-car &  car \\
    &product-food &  food \\ 
    &product-game & game \\
    &product-other &  other product \\ 
    &product-ship & ship \\
    &product-software &  software \\
    &product-train &  train \\
    &product-weapon &  weapon \\
    &other-astronomything & astronomy thing \\
    &other-award &  award \\
    &other-biologything &  biology thing \\ 
    &other-chemicalthing & chemical thing \\ 
    &other-currency & currency \\
    &other-disease &  disease \\
    &other-educationaldegree &  educational degree \\ 
    &other-god & god \\ 
    &other-language & language \\
    &other-law &  law \\
    &other-livingthing & living thing \\
    &other-medical & medical \\

    &building-airport & airport \\
    &building-hospital & hospital \\
    &building-hotel & hotel \\
    &building-library & library \\
    &building-other & other building \\
    &building-restaurant & restaurant \\
    &building-sportsfacility & sports facility \\
    &building-theater & theater \\
    &\makecell[c]{event-attack/battle\\/war/militaryconflict} & \makecell[l]{attack battle \\war military conflict} \\
    &event-disaster & disaster \\
    &event-election & election \\
    &event-other & other event \\
    &event-protest & protest \\
    &event-sportsevent & sports event \\
    &location-bodiesofwater & bodies of water \\
    &location-GPE & \makecell[l]{geographical social \\political entity} \\
    &location-island & island \\
    &location-mountain & mountain \\
    &location-other & other location \\
    &location-park & park \\
    &\makecell[c]{location-road/railway\\/highway/transit} & \makecell[l]{road railway \\highway transit} \\
    &organization-company & company \\
    &organization-education & education \\
    &\makecell[c]{organization-government\\/governmentagency} & government agency \\
    &organization-media/newspaper & media newspaper \\
    &organization-other & other organization \\
    &organization-politicalparty & political party \\
    &organization-religion & religion \\
    &organization-showorganization & show organization \\
    &organization-sportsleague & sports league \\
    &organization-sportsteam & sports team \\

\bottomrule
\end{tabular}

}

\end{center}
\caption{Original labels and their corresponding natural-language-form type names of Few-NERD.}
\label{tab:dataset_labels_nlf_1}
\end{table}

\begin{table}[htb]
\small
\begin{center}
\resizebox{\columnwidth}{!} {
\begin{tabular}{lcl}
\toprule
\bf Dataset & \makecell[c]{\textbf{Labels}} & \makecell[c]{\textbf{Type names}} \\

\cmidrule(lr){1-1} \cmidrule(lr){2-3}
\multirow{23}{*}{\textbf{I2B2'14} } 
& AGE & age \\
& BIOID & biometric ID \\
& CITY & city \\
& COUNTRY & country \\
& DATE & date \\
& DEVICE & device \\
& DOCTOR & doctor \\
& EMAIL & email \\
& FAX & fax \\
& HEALTHPLAN & health plan number \\
& HOSPITAL & hospital \\
& IDNUM & ID number \\
& LOCATION\_OTHER & location \\
& MEDICALRECORD & medical record \\
& ORGANIZATION & organization \\
& PATIENT & patient \\
& PHONE & phone number \\
& PROFESSION & profession \\
& STATE & state \\
& STREET & street \\
& URL & url \\
& USERNAME & username \\
& ZIP & zip code \\

\cmidrule(lr){1-1} \cmidrule(lr){2-3}
\multirow{4}{*}{\textbf{CoNLL'03} } 
& PER & person \\
& LOC & location \\
& ORG & organization \\
& MISC & miscellaneous \\

\cmidrule(lr){1-1} \cmidrule(lr){2-3}
\multirow{11}{*}{\textbf{GUM} } 
& abstract & abstract \\
& animal & animal \\
& event & event \\
& object & object \\
& organization & organization \\
& person & person \\
& place & place \\
& plant & plant \\
& quantity & quantity \\
& substance & substance \\
& time & time \\

\cmidrule(lr){1-1} \cmidrule(lr){2-3}
\multirow{6}{*}{\textbf{WNUT'17} } 
& corporation & corporation \\
& creative-work & creative work \\
& group & group \\
& location & location \\
& person & person \\
& product & product \\

\cmidrule(lr){1-1} \cmidrule(lr){2-3}
\multirow{18}{*}{\textbf{Ontonotes} } 
& CARDINAL & cardinal \\
& DATE & date \\
& EVENT & event \\
& FAC & fac \\
& GPE & \makecell[l]{geographical social \\political entity} \\
& LANGUAGE & language \\
& LAW & law \\
& LOC & location \\
& MONEY & money \\
& NORP & nationality religion \\
& ORDINAL & ordinal \\
& ORG & organization \\
& PERCENT & percent \\
& PERSON & person \\
& PRODUCT & product \\
& QUANTITY & quantity \\
& TIME & time \\
& WORK\_OF\_ART & work of art \\

\bottomrule
\end{tabular}

}

\end{center}
\caption{Original labels and their corresponding natural-language-form type names of datasets under Cross-Dataset settings.}
\label{tab:dataset_labels_nlf_2}
\end{table}

\end{document}